\documentclass{article}
\newtheorem{lem}{Lemma}[section]
\newtheorem{tr}{Theorem}[section]
\def\b{B}

\begin{document}

\title{A Topological Code for Plane Images}
\author{Evgeny V. Shchepin}
\thanks{This work is supported by the Russian Government project 11.G34.31.0053.
and Russian Foundation of Basic Researches grant 11-01-00822}

\maketitle

\begin{abstract}{It is proposed a new code
for contours of black-white plane images. This code was applied
for optical character recognition of printed and
handwritten characters. One can apply it to recognition
of any visual images.}
\end{abstract}
\bigskip
\bigskip

{\it Address:}  Russia, 150000, Yaroslavl, Sovetskaya str.
14, Yaroslavl State University, Delone Laboratory of Discrete and
Computational Geometry.
\bigskip

{\it e-mail:} scepin@mi.ras.ru

\bigskip
\bigskip

\bigskip

\section{Introduction}

A general approach to character recognition via
critical points was presented in joint papers of
the author with G. Nepomnyashchii \cite{NS},\cite{SN}
in 1991. These papers present a theoretical background
of the critical point approach and inform on the first results
obtained in its implementation.
But these papers do not present the concrete code
which was realized by programming. The reason is that
this code was a commercial secret in that time.
But now the author has finished his
software business and has no more obstacles to open the code to the public.
Let us call this code CRIPT-code, this was the name of
OCR-programs which was produced under the
direction of the author in the years 1990-1995  in some
small russian software firm named "Scriptum".

\section{Topological background}

The following exposition one can consider as a 1-dimensional
version of the Morse theory \cite{M}. This theory is very
elementary. That is why we omit some simple proofs.
For practical work one does need proofs, that is the
reason to omit a long proofs.

We always suppose that on plane is fixed a system of Cartesian
coordinates. The second coordinate will be called height.
The set of all points of the plane which has a fixed height $h$
will be called a {\it $h$-level line} or {\it $h$-level}.
Each $h$-level line divides plane into two semiplanes
which we will call {\it $h$-upper} and
{\it $h$-lower}.
If $H>h$ the intersection of $H$-lower and $h$-upper semiplanes
is called horizontal $[h,H]$-band.

\subsection{Height classification of arcs}
The reader can understand the word {\it arc} as  a
{\it broken line without self-intersection}, although
almost all consideration are valid for an arbitrary Jordan curve.
{\it Closed arc} is an arc without ends and nonclosed one has two ends.
The projection of an arc on the second coordinate is said to
be {\it height range} of the arc. The heights of the ends
of an arc will be called {\it ends levels}.
The arcs which are of interest are connectivity components
of intersection of the boundary of some plane domain with
a horizontal band. The ends of such arcs belongs to the
boundary of the band. This motivates the following definition.

An arc will be called {\it $[h,H]$-bounded} if it is contained in
the $[h,H]$-band and both its  ends lyes on the boundary
of the band (i.e. on the levels $h$ and $H$).

We will divide $[h,H]$-bounded arcs into four height's types
corresponding to letters $\b$, $C$, $D$, $O$.

The type $C$ is associated with the word "Crossing".
An arc $\alpha$ has the type $C$
or that it is a {\it crossing} if the ends of its height's range
coincides with its ends levels.

And we will call an arc $\alpha$
as {\it $[h,H]$-crossing} if it is a crossing with
the height range $[h,H]$.

The type $\b$ is associated with the words "Birth".
An arc $\alpha$ has the type $\b$ if it is contained in $h$-upper
semiplane and both its ends have the same height.

The type $D$ is associated with the words "Death".
An arc $\alpha$ has the type $D$ if it is contained in the $h$-lower
semiplane and both its ends have the same height.

And finally an arc is called to have the type $O$ if it is
an closed arc because closed arcs are $O$-like.

\subsection {Critical levels}

Let us say that an arc $\alpha$ {\it touches} a $h$-level  if
$h$ is the end of the height range of $\alpha$ and is not
its ends level. In the case
$\alpha$ is contained in the $h$-upper
(upper touching)  or $h$-lower (lower touching) semiplane.

A number $h$ is called a {\it critical height} of an arc $\alpha$
and
$h$-level is called {\it $\alpha$-critical} if there is
such subarc $\alpha'\subset \alpha$, that touches $h$-level.

An arc $\alpha$ without critical heights is are called {\it uncritical}.

The proof of the following lemmas presents
a simple exercise on definitions.
\begin{lem}\label{TD} Every uncritical arc is a crossing.
\end{lem}
\begin{lem}\label{Monot} Suppose that an $[h,H]$-bounded arc $\alpha$
has at most one critical level $g$. In the case the intersection
of $\alpha$ with the $g$-level is connected.
\end{lem}

\subsection{Crossing decomposition}

Throughout the section we consider a bounded
domain denoted by $S$
with boundary  denoted  by $\partial S$.
We do not suppose $S$ to be connected. It may contain
a finite number of connectivity components.
We will suppose that the boundary is  piecewise linear.

A horizontal $[h,H]$-band is called {\it regular} with respect
to the domain $S$ (shortly $S$-regular)
if $h$ and $H$ are not $\partial S$-critical.
In the sequel we will consider only $S$-regular bands.
The intersection of  $\partial S$ with
the horizontal $[h,H]$-band is denoted  $\partial S[h,H]$.
The components of the intersection $\partial S[h,H]$.
are $[h,H]$-arcs.

Let us say that a subarc $\alpha'$ of an arc $\alpha$ is
its {\it crossing component} if $\alpha'$ is crossing and any
subarc of $\alpha$ containing $\alpha'$ is not crossing.

\begin{lem} If an arc $\alpha$ has type $\b$, $D$
or $O$ it contains two crossing components,
these components have the same range as $\alpha$ and
its union coincides with $\alpha$.
\end{lem}

For every crossing component $\alpha$ of
$\partial S[h,H]$ let us introduce two numbers.
The first one we denote by $L(\alpha)$ and name {\it lower position},
this number is equal to $0$ if $\alpha$ has not an end on the level $h$
and is equal to the number of crossing component of $\partial S[h,H]$
which have an $h$-level end  which lies on the left of the $h$-level end
of the $\alpha$.
The second number is denoted by $U(\alpha)$ named as
{\it upper position}, it is defined as $0$ if $\alpha$ has not an
$H$-level end and is defined as
the number of crossing component of $\partial S[h,H]$ which have an
$H$-level end  on the left of the $H$-level end of $\alpha$.

\begin{lem} \label{Sum} Let $h_1<h_2<h_3$ be three $S$-uncritical levels.
Let $\alpha$ be a crossing component of $\partial S[h_1,h_2]$ with $U(\alpha)>0$
 and
$\beta$ be a crossing component of $\partial S[h_2,h_3]$.
The intersection $\alpha\cap\beta$ is nonempty iff $U(\alpha)=L(\beta)$.
In this case $\alpha\cup\beta$ is a crossing component of
$\partial S[h_1,h_3]$
\end{lem}

\subsection{CRIPT-code of the domain}

Let us say that  a $[h,H]$-band is  {\it $S$-unicritical}
if it is $S$-uncritical or
if the segment $[h,H]$ contains only one $S$-critical height
in its interior.
Let us remark that for this case no connectivity component of
$\partial S[h,H]$ has the type $O$.

Let $\alpha$ and $\beta$ be two disjoint crossing component of $\partial S[h,H]$.
Let us say that $\alpha$ is {\it on the left from} $\beta$ if the intersection
of $\alpha$ with the critical level is on the left of the intersection
of $\beta$ with the critical level. (By virtue of the lemma \ref{Monot}
both intersection are points or intervals, hence the relation
"on the left from" is defined correctly).
If $\alpha$ and $\beta$ are intersecting crossing components of
$\partial S[h,H]$ then its union is an $[h,H]$-arc of the
type $B$ or $D$.  If the type is $B$ let us say that $\alpha$ is
on the left from $\beta$ if $L(\alpha)<L(\beta)$.
If the type is $D$ let us say that $\alpha$ is on the left of
$\beta$ if $U(\alpha)<U(\beta)$.

For an $S$-unicritical $[h,H]$-band and for every crossing component
$\alpha$ of $\partial S[h,H]$ one define its {\it position} $P(\alpha)$ as
the number of crossing components of $\partial S[h,H]$ which
are on the left from $\alpha$.

To define the CRIPT-code of the $\partial S[h,H]$ let us consider
the sequence $\alpha_1,\alpha_2,\dots A_n$ of crossing components of $\partial S[h,H]$
ordered by increasing of the position.

For every crossing component $\alpha_i$ let us denote by $T(A_i)$
the type of connectivity component of $\partial S[h,H]$  to
which it belongs.

The sequence $T(\alpha_1),T(\alpha_2),\dots,T(\alpha_n)$ is called
is called  {\it $[h,H]$-CRIPT-string} of the domain $S$.

CRIPT-string which contains the only type $C$ is called
{\it trivial}. The trivial CRIPT-string one obtains if
the band is uncritical.

Let us say that an increasing sequence of numbers $h_1, h_2,\dots h_m$
is an {\it unicritical decomposition} of $S$,
if $S$ is contained  in the band $[h_1,h_m]$ and all bands
$[h_i,h_{i+1}]$ are unicritical.

For any unicritical decomposition $h_1, h_2,\dots h_m$ of
$S$ let us define $[h_1, h_2,\dots h_m]$-CRIPT code of $S$
as the sequence of $[h_i,h_{i+1}]$-CRIPT strings separated
by $;$.

An unicritical decomposition $h_1, h_2,\dots h_m$ of $S$
with minimal possible $m$ is called {\it minimal decomposition}.
The CRIPT-code of a minimal decomposition of $S$ is
called  {\it minimal CRIPT-code} of $S$. This code
does not depends of the choice of a minimal decomposition
and is uniquely defined by the domain.
One can obtain the minimal CRIPT-code using
any unicritical decomposition $h_1, h_2,\dots h_m$
simply eliminating  in the $[h_1, h_2,\dots h_m]$-CRIPT code
trivial strings. The proof of these facts easily follows
from the following lemma.

\begin{lem} Let two bands $[h_1,H_1]$ and $[h_2,H_2]$
be both $S$-unicritical and contain the same
$S$-critical level.
In this case $[h_1,H_1]$-CRIPT-string and $[h_2,H_2]$-CRIPT-string
coincide.
\end{lem}
Proof.

Let us consider at first the case $h_1=h_2=h$
and $H_1<H_2$. The band $[H_1,H_2]$ does not contain
$S$-critical levels, and by virtue of lemma \ref{TD}
the intersection of $\partial S$ with the band consists only of
$[H_1,H_2]$-crossings.
For every end $a$ of a connectivity component $\alpha$ of
$\partial S[h,H_1]$
on the level $H_1$ there is unique $[H_1,H_2]$-crossing
which contains $a$. Adding to  $\alpha$ this crossing
one obtains an connectivity component of $\partial S[h,H_2]$, which
has the same type. And it is obvious that such operation
conserve the position. Hence $[h,H_1]$-CRIPT-string
coincides with $[h,H_2]$-CRIPT-string.

If $h_1>h_2$ the above arguments prove coincidence of CRIPT-strings
for bands $[h_1,H_2]$ and $[h_1,H_1]$. The coincidence of CRIPT-strings
for bands $[h_1,H_2]$ and $[h_2,H_2]$ can be proved by the same
arguments. The lemma is proved.

The above lemma justifies the following definition.
Let $h$ be a critical level for $\partial S$, let us define
$h$-CRIPT-string of the domain $S$  as $[H_1,H_2]$-CRIPT-string
for some unicritical band $[H_1,H_2]$ containing $h$.

As easy to see the minimal CRIPT-code of $S$
coincides with
the sequence
of its CRIPT-strings corresponding to all its critical levels
taken in the order of decreasing of level.

The CRIPT-code of a domain can be
represented as a word of the alphabet $\b,C,D,;$.
For example the letters $A$, $\b$, $O$ has the following
CRIPT-codes \footnote{The code of a letter is the code of some
its small neighborhood}

\b\b;C\b\b C;CDDC;C\b\b C;DDDD (letter $A$)

\b\b ;C\b\b C;CDDC;C\b\b C;CDDC;DD (letter $B$)

\b\b ;C\b\b C;CDDC;DD (letter $O$).

One can prove the following theorem which
justifies the application of minimal CRIPT-code for
recognition.
\begin{tr}Two domains has the same minimal CRIPT-code iff there exist
such preserving orientation homeomorphism of its interiors which
preserves height-levels, i.e. if two points have the same height
its images have  the same property.
\end{tr}

\section{Bit-Maps Coding}
Theoretically to produce the CRIPT-code of a digital picture,
which has the form of a bit-map it is sufficiently to indicate
a way how to associate with it a plane region and then to consider
the CRIPT-code of its boundary.
This construction may be performed in the following way:
at first one assigns to the bit-map the finite set of
points with integral coordinates, the first coordinate is the
number of the column of a black pixel (black corresponds to 1,
and white corresponds to 0 presents the background
of the picture), the second is the number of
its string. At second one assigns to each such integral point
the rectangle which has it as its center of symmetry and has
sides parallel to coordinates, such that its height is equal to 1
and its width is equal 1.1.
Now the associated domain can be defined as the union of
the rectangles associated with black pixels.
The CRIPT-code of a bit-map is defined as the CRIPT-code
of associated domain.

In practice the generation of CRIPT-code is performed during the first
scanning of the bit-map image simultaneously with dividing of
the image into connectivity components, which is a necessary
operation for any OCR-program.

Let us remark that all critical levels of the associated
domain has fractional part $1/2$.
Therefore all bands of the type $[k,k+1]$ for a natural $k$
are  unicritical and one can apply it to
calculate the CRIPT-code of the associated domain.
To calculate a CRIPT-string corresponding to level $k+1/2$
one has to consider $k$-th and $(k+1)$-th strings of the bitmap.
The practical algorithm produces from a string
of a bit-map the switching sequence which represent the
sequence of natural numbers $n_1,n_2,\dots n_k$ which correspond
to positions of switching from white to black and from black to
white. So $n_1$ is the number of position of the first (from the left)
black pixel, and $n_k-1$ is the number of position of the last black
pixel. If  $n_1,n_2,\dots n_k$ is the switching sequence of
$k$-th string and  $m_1,m_2,\dots m_l$ is the switching sequence of
the next string we can construct CRIPT-code for this pair, which
corresponds to the level $(k+1/2)$.
Namely, all pairs of $\b$-codes of considered band are in one-to-one correspondence
with such 4-tuples $m_i,m_{i+1},n_j,n_{j+1}$ which satisfy one
of the following
conditions:
\begin{enumerate}
\item  if $i$ is odd then $j$ is even and $n_j<m_i<m_{i+1}<n_{j+1}$
\item  if $i$ is even then $j$ is odd and $n_j\le m_i<m_{i+1}\le n_{j+1}$
\end{enumerate}
Dually, all pairs of $D$-codes correspond to 4-tuples
$m_i,m_{i+1},n_j,n_{j+1}$ which satisfy one of the following
conditions:
\begin{enumerate}
\item  if $j$ is odd then $i$ is even and $m_i<n_j<n_{j+1}<m_{i+1}$
\item  if $j$ is even then $i$ is odd and $m_i\le n_j<n_{j+1}\le m_{i+1}$.
\end{enumerate}
All $C$ codes corresponds to such pairs $n_j,m_i$ that
\begin{enumerate}
\item $i+j$ is even and $n_j=m_i$
\item if $i$ is odd then $j$ is odd
$m_{i-1}<n_j<m_i\le n_{j+1}$ or $n_{j-1}<m_i<n_{j}\le m_{i+1}$
\item if $i$ is even then $j$ is even
$m_{i-1}\le n_j<m_i< n_{j+1}$ or $n_{j-1}\le m_i<n_{j}< m_{i+1}$
\end{enumerate}

\section{Structure of  CRIPT-code}

The CRIPT-code of a domain $S$ represent a word of
the CRIPT-alphabet $\{B,C,D,;\}$. Let us say that a word
of CRIPT-alphabet is {\it correct} if there is such a
domain which has this word as its CRIPT-code.
Let us find conditions which characterize the correctness.
To do it we have to introduce some definitions.
For any word $w$ by $w[k,n]$ we denote its subword
which start with $k$-th letter of $w$ and terminates
at the $n$-th letter of it. So $w[k,k]$ is
the $k$-th letter of $w$. If $k>n$ we pose $w[k,n]$ is the empty word.
Let us denote by $B(w)$, $C(w)$, $D(w)$, $E(w)$
correspondingly the number of letters  $B$, $C$, $D$ and $;$ in it.
So $E(c)$ is the number of strings in a CRIPT-code.
Supposing that $E(w)=-1$ for empty word one
can define the $k$-th string of any word $w$
as $w[n,m]$, where $E(w[1,n-2])<E(w[1,n-1])=E(w[1,m])<E(w[1,m+1])=k$.
The notation for the $k$-th string of $w$ is $w[k]$.

The first obvious condition is that the first string
 of a correct word contains only letters $B$.
One can write this condition as $B(w[1])>0$, $C(w[1])=0$
and $D(w[1])=0$.
 The dual
condition is that the last string contains only
letters $D$. Let us name this conditions
{\it boundary conditions}.

The second pair of conditions we name {\it evenness conditions}.
If before a symbol of the code of the type $B$ or $D$
there are even number of the same symbols then the next
symbol of the code has the same type.
For the letter $B$ the evenness condition one can wright
as follows: if $B(w[1,n])-1=B(w[1,n-1])$ is an even number
then $B(w[1,n+1])=B(w[1,n])+1$.

Let us say that two symbols of the same type $B$ or $D$
are conjugated to each other if before the first of them
one has even number of the same symbols and the second is
the succeeder of the first one.
Hence in a correct code all symbols of types $B$ and $D$
are divided into pairs of conjugated symbols.
Geometrically conjugated symbols represent pairs of
crossing component of the same connectivity component.

The third type of conditions we will name {\it balance conditions}.
Let $h_{k-1}$, $h_k$ and $h_{k+1}$ denotes such levels that
the bands $[h_{k-1},h_k]$ and $[h_{k},h_{k+1}]$ are $S$-unicritical
and its CRIPT-strings represent $k$-th and $(k+1)$-th strings
of the CRIPT-code $w$ of $S$.

The number of component of intersection of $\partial S$ with
the level $h$ from one side is equal to $B(w[k])+C(w[k])$
because the last number is the number of ends of crossing
component of $\partial S[h_k,h_{k+1}]$. And from the other
side it is equal to $D(w[k+1])+C(w[k+1])$ because this number
is equal to the number of ends of crossing components of
 $\partial S[h_{k-1},h_{k}]$.
As result one obtains for every $k$ the following
{\it balance equality}:
$$B(w[k])+C(w[k])=D(w[k+1])+C(w[k+1])$$

Summing the balance equalities one obtains the following
global equality $B(w)=D(w)$ for every correct word $w$.

\subsection{Contactness}
Let $v$ and $w$ are two one-string words ($E(v)=E(w)=1)$)
which satisfy the balance equality
$B(v)+C(v)=D(w)+C(w)$. For such strings we introduce
the following {\it  contactness} relation.
We will say that a letter  of $v$ on $k$-th position
contacts with a letter of $w$ on the $n$-th position
if the type of the first letter is $B$ or $C$, the type
of the second letter is $C$ or $D$ and
$B(v[1,k-1])+C(v[1,k-1])=C(w[1,n-1])+D(w[1,n-1])$.
The geometrical meaning of the contactness was demonstrated
by the lemma \ref{Sum}. The contacting codes
for CRIPT-strings of adjancent unicritical bands corresponds
to intersecting crossing components.

Now we are ready to outline the proof of the following {\it realization
theorem}.
\begin{tr} Every word $w$ of the CRIPT-alphabet
which satisfies the boundary conditions, the evenness
conditions and balance conditions is correct.
\end{tr}
Proof.

To construct the domain which has the given word $w$
as its CRIPT-code let us chose an increasing sequence
$h_1,h_2,\dots ,h_n$, where $n=E(w)$.
On every $h_i$-level let us choose the sequence
$P_{i,1},P_{i,2},\dots,P_{i,k(i)}$ of points
where $k(i)$ is the number of letters in $w[i]$.
Suppose that these points are ordered by the increasing
of the first coordinate.

Now let us join by right segments all pairs of points
corresponding to conjugated codes and join all pairs
of contacted points. We omit the proof that the
constructed  domain
has the CRIPT-code equal to $w$.

From the above considerations one can get an algorithm
of extracting of CRIPT-codes of connectivity components
from the CRIPT-code of an disconnected domain.
Indeed, two letters of the CRIPT-code of the domain
belongs to the same connectivity component iff
there is a sequence of letters of the code such that
the first term of the sequence coincide with the
first given letter, the last one coincide with the
second given letter and for any letter of the sequence
its succeeder is or conjugated or contacted to it.

\subsection{Elimination of distortions}
And another practical problem is elimination of
distortions. To do it one can clusterize the critical
levels and separate clusters to generate the CRIPT-code.
In this case one has to work with CRIPT-strings
of bands which are not unicritical. For this
bands can arise arcs of type $O$.
One can encounter a difficulties to order crossing
components in the case because
some crossing component can have disjoint height
ranges.
But all these difficulties have a practical solutions.


\end{document}